\documentclass[sigconf]{acmart}

\AtBeginDocument{%
  }

\copyrightyear{2026}
\acmYear{2026}
\setcopyright{cc}
\setcctype{by}
\acmConference[CIKM '26]{Proceedings of the 35th ACM International Conference on Information and Knowledge Management}{November 07--11, 2026}{Rome, Italy}
\acmBooktitle{Proceedings of the 35th ACM International Conference on Information and Knowledge Management (CIKM '26), November 07--11, 2026, Rome, Italy}
\acmDOI{10.1145/3799682.3840048}
\acmISBN{979-8-4007-2539-5/2026/11}

\usepackage{booktabs}
\usepackage{amsmath}
\usepackage{enumitem}
\usepackage{caption}
\usepackage{algorithm}
\usepackage{algpseudocode}
\usepackage{colortbl}
\usepackage{balance}
\definecolor{methodrow}{RGB}{222, 235, 247}

\newcommand{\method}{LatentRec}
\newcommand{\hk}{\mathbf{h}_k}
\newcommand{\hzero}{\mathbf{h}_0}
\newcommand{\Rspace}{\mathbb{R}}

\begin{document}

\title{Do Recommenders Need to Think in Words? Latent Reasoning for Efficient LLM-Based Recommendation}

\author{Hongyang Su}
\orcid{0000-0003-3763-0703}
\email{supersusu@tencent.com}
\affiliation{%
  \institution{Tencent}
  \city{Shenzhen}
  \state{Guangdong}
  \country{China}
}

\author{Beibei Kong}
\authornote{Corresponding author.}
\orcid{0009-0001-9075-824X}
\email{echokong@tencent.com}
\affiliation{%
  \institution{Tencent}
  \city{Shenzhen}
  \state{Guangdong}
  \country{China}
}

\author{Lei Cheng}
\orcid{0000-0001-7681-9758}
\email{raycheng@tencent.com}
\affiliation{%
  \institution{Tencent}
  \city{Shenzhen}
  \state{Guangdong}
  \country{China}
}

\author{Chengxiang Zhuo}
\orcid{0009-0001-5750-140X}
\email{felixzhuo@tencent.com}
\affiliation{%
  \institution{Tencent}
  \city{Shenzhen}
  \state{Guangdong}
  \country{China}
}

\author{Zang Li}
\orcid{0000-0002-2305-7179}
\email{gavinzli@tencent.com}
\affiliation{%
  \institution{Tencent}
  \city{Shenzhen}
  \state{Guangdong}
  \country{China}
}

\renewcommand{\shortauthors}{Hongyang Su, Beibei Kong, Lei Cheng, Chengxiang Zhuo, and Zang Li}

\begin{abstract}
LLM-based recommender systems increasingly generate chain-of-thought (CoT) text before each item prediction to activate the LLM's reasoning capacity, costing tens to hundreds of extra tokens per recommendation at inference. We ask whether inference-time reasoning artifacts (textual or latent) can be replaced by training-time supervision alone, without loss of accuracy.
We propose \method{}, a framework for LLM-based generative recommendation that replaces inference-time reasoning artifacts with training-time per-step head supervision. \method{} has three components. A 2.5M-parameter Continuous Thought Module iterates $K$ latent reasoning steps over the LLM's hidden state during training. Soft Token Selection bridges the $\sim$14$\times$ norm gap between hidden states and input embeddings. A Per-step Reasoning Loss (PRL) supervises every intermediate state through the LLM's prediction head against the first token of the target Semantic ID. The trained LLM retains this accuracy without the module at inference, so the module can be discarded at deployment.
On three Amazon benchmarks, \method{} improves HR@10 by $4.0$--$5.0\%$ (relative) over a CoT-distilled baseline at $18.3\times$ lower latency (batch$=$1). Per-step training supervision consistently improves accuracy across three datasets, paid once at training rather than at every recommendation.
\end{abstract}

\ccsdesc[500]{Information systems~Recommender systems}
\ccsdesc[300]{Computing methodologies~Natural language processing}

\keywords{Recommender Systems, Large Language Models, Latent Reasoning, Chain-of-Thought}

\maketitle

\section{Introduction}
\label{sec:intro}

Recent LLM-based recommenders~\cite{llm4rec-survey-fan, tallrec} increasingly bind reasoning to inference-time generation: chain-of-thought (CoT)~\cite{wei2022chain} trajectories are written as language tokens before each item prediction, exemplified by CoT distillation~\cite{onerec-think} and related variants~\cite{thinkrec, r2ec, reczero}. Latent reasoning relaxes the textual constraint by replacing inference-time text with continuous hidden-state computation, established for math and logic by COCONUT~\cite{coconut}, CODI~\cite{codi}, and pause-token continuation~\cite{pause-tokens}, and recently brought to recommendation through encoder-side rollouts~\cite{rearec, lares, plr} and concurrent RL-based latent reasoning for LLM recommenders~\cite{latentr3}. Across both regimes, however, reasoning capacity remains whatever the model produces at inference, paid for at every recommendation request.

\emph{Can inference-time reasoning artifacts (textual or latent) be replaced by training-time supervision alone, without loss of accuracy?}
We propose \method{}, a framework for SID-generative LLM recommenders that replaces inference-time reasoning with training-time per-step head supervision. At the heart of \method{} is a Per-step Reasoning Loss (PRL), inspired by progressive supervision~\cite{rearec}, that supervises every intermediate hidden state through the LLM's prediction head against the first token of the target Semantic ID. PRL needs intermediate states to supervise, which a lightweight Continuous Thought Module (2.5M parameters, $<$0.2\% of the LLM) generates by iterating $K$ latent reasoning steps over the LLM's hidden state during training, with Soft Token Selection projecting each state back into the input-embedding distribution the LLM was pre-trained on. The trained LLM retains the resulting accuracy without the module at inference, so the Continuous Thought Module is required only during training and can be discarded at deployment.

\method{} differs from prior latent reasoning recommenders in two respects.
(i)~It operates in the SID-generative regime where items are decoded as multi-token Semantic IDs. This regime exhibits a $\sim$14$\times$ norm gap between LLM hidden states and input embeddings (\S\ref{sec:sts}), absent in encoder-side rollouts~\cite{rearec, lares, plr} or title-decoded LLM recommenders~\cite{latentr3}.
(ii)~Its per-step prediction-head supervision adapts the progressive supervision idea of~\cite{rearec} to the LLM SID-decoder regime, unlike objectives that supervise only the final SID or optimize final rewards over reasoning traces.

Our contributions are summarized as follows.
(1)~\textbf{Paradigm shift.}~We propose \method{}, which replaces inference-time reasoning artifacts (textual or latent) in SID-generative LLM recommenders with training-time per-step supervision, eliminating per-request reasoning cost without sacrificing accuracy.
(2)~\textbf{Technical contribution.}~We adapt progressive supervision to the LLM SID-decoder regime through a Per-step Reasoning Loss applied via the shared prediction head, paired with Soft Token Selection that resolves the representation-space mismatch unique to LLM SID-decoding. The added module is 2.5M parameters ($<$0.2\% of the LLM) and is discarded at deployment.
(3)~\textbf{Empirical impact.}~On three Amazon benchmarks, \method{} improves HR@10 by $4.0$--$5.0\%$ (relative) over a same-backbone CoT-distilled baseline at $18.3\times$ lower inference latency, moving the cost of reasoning from per-request inference to one-time training.
(4)~\textbf{Mechanistic insight.}~We isolate where chain-of-thought actually contributes: per-step supervision drives the gain at $K{=}0$ inference, while CoT pre-training mainly stabilizes deeper latent rollout (without it, accuracy drops $11\%$ at $K{=}16$). Multi-step latent computation accrues value at training rather than at inference.

\section{Method}
\label{sec:method}

\begin{figure}[!t]
    \centering
    \includegraphics[width=\linewidth]{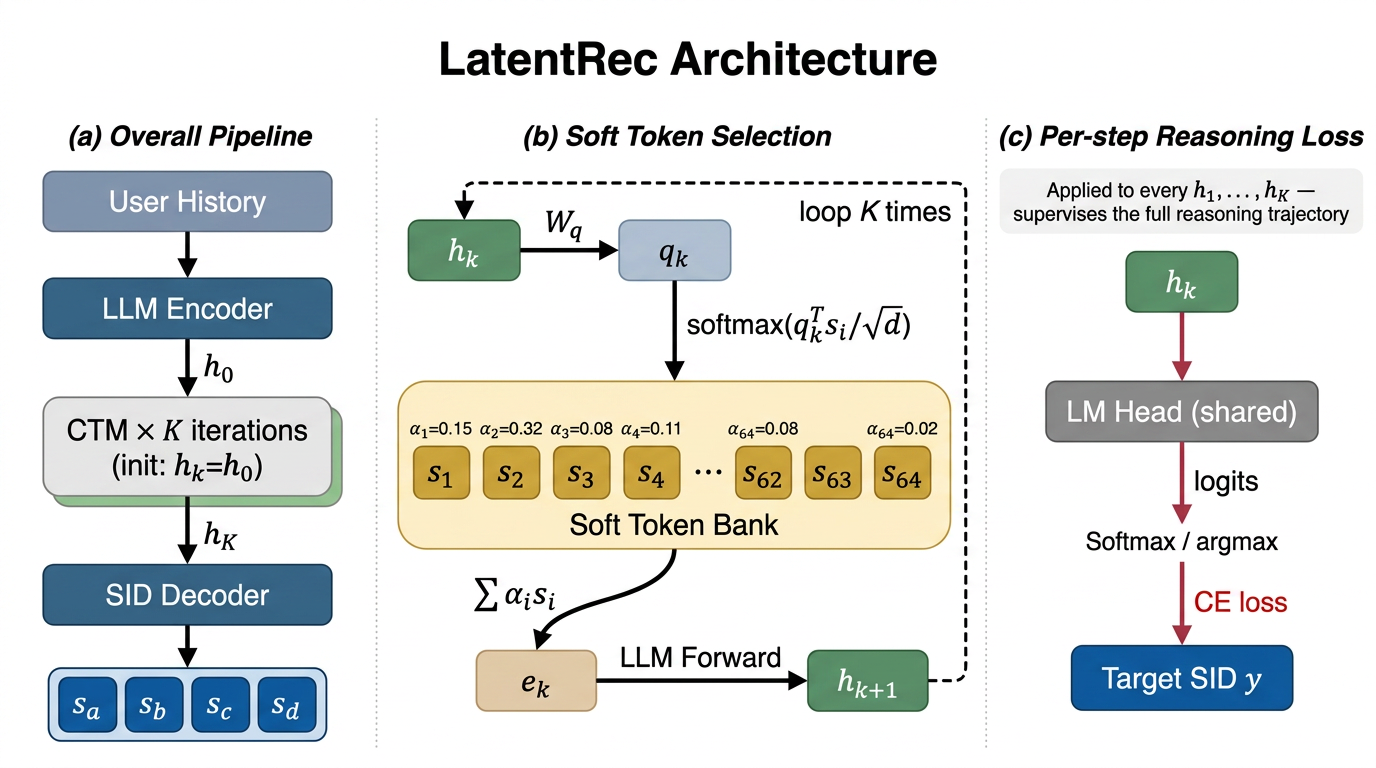}
    \caption{\method{} architecture. (a) Overall pipeline. (b) Soft Token Selection projects each hidden state into the LLM's input-embedding distribution. (c) Per-step Reasoning Loss supervises every intermediate state via the shared LM head against the first SID component of the target.}
    \Description{Architecture diagram of LatentRec illustrating the Continuous Thought Module performing K iterative steps over the LLM hidden state, the Soft Token Selection projection, and Per-step Reasoning Loss applied to each intermediate state.}
    \label{fig:arch}
\end{figure}

\method{} replaces inference-time reasoning with training-time per-step head supervision for LLM-based generative recommendation, applicable to any LLM that decodes items as hierarchical Semantic IDs (SIDs)~\cite{tiger,minionerec,onerec-think}.
Three components work together (Figure~\ref{fig:arch}): (1)~a Continuous Thought Module that performs hidden-state iterations during training, (2)~Soft Token Selection that bridges the representation space gap between hidden states and input embeddings, and (3)~Per-step Reasoning Loss (PRL) that supervises every intermediate state through the LLM's prediction head against the first component of the target Semantic ID.

\subsection{Continuous Thought Module}
\label{sec:ctm}

The Continuous Thought Module produces the intermediate hidden states $\mathbf{h}_1, \ldots, \mathbf{h}_K$ that PRL supervises. Starting from the user-history hidden state $\hzero \in \Rspace^d$, $K$ reasoning steps unroll as
\begin{equation}
    \mathbf{e}_k = \phi(\hk), \quad \mathbf{h}_{k{+}1} = \text{LLM}(\mathbf{e}_k, \text{KV}^{(k)})
    \label{eq:iteration}
\end{equation}
where $\phi$ is Soft Token Selection (\S\ref{sec:sts}) and $\text{KV}^{(k)}$ is the accumulated key-value cache (Algorithm~\ref{alg:train}, lines~2--5). The 2.5M trainable parameters added by \method{} are confined to STS ($\mathbf{S}$ and $W_q$). Each step nevertheless invokes the full LLM, so latent computation cost scales as $K \times P$ where $P$ is one LLM forward pass. At inference $K{=}0$ disables the rollout and decodes directly from $\hzero$.

\subsection{Soft Token Selection}
\label{sec:sts}

A key challenge is the \emph{representation space mismatch}: in our Qwen3-1.7B backbone, the mean L2 norm of last-layer hidden states ($\|\mathbf{h}\| \approx 45$) is roughly $14\times$ that of input embeddings ($\|\mathbf{e}\| \approx 3$).
Prior latent reasoning methods in math and logic domains~\cite{coconut,codi} feed raw hidden states back as inputs and rely on careful training stabilization.
The LLM-based SID-decoding setting amplifies this challenge: every intermediate state $\mathbf{h}_k$ is supervised through the LLM's prediction head (\S\ref{sec:prl}) and must align with the input-embedding distribution the LLM was pre-trained on, otherwise the per-step logits drift away from the SID vocabulary as $K$ grows.

We maintain $N$ learnable soft tokens $\mathbf{S} \in \Rspace^{N \times d}$, initialized from the LLM's vocabulary.
The projection computes:
\begin{equation}
    \alpha_i^{(k)} = \frac{\exp(\mathbf{q}_k^\top \mathbf{s}_i / \sqrt{d})}{\sum_j \exp(\mathbf{q}_k^\top \mathbf{s}_j / \sqrt{d})}, \quad
    \mathbf{e}_k = \sum_{i} \alpha_i^{(k)} \mathbf{s}_i
    \label{eq:sts_eq}
\end{equation}
where $\mathbf{q}_k = W_q \hk$.
The convex combination keeps the projected output within the input-embedding scale, bounds gradients by $\alpha_i \in (0,1)$, and avoids a separate normalization layer. Beyond matching scale, the convex combination keeps every projected $\mathbf{e}_k$ in-distribution with respect to the LLM's input-embedding manifold, both in norm and in direction. A scalar rescaling would match scale but not direction, while raw hidden-state feedback as used in math/logic latent reasoning~\cite{coconut, codi} matches neither. Empirically, combining raw hidden-state feedback with our per-step head supervision (\S\ref{sec:prl}) diverges within the first epoch under matched configuration, with per-step logits drifting off the SID vocabulary and triggering our auto-stop guard ($\mathcal{L} > 200$). The failure mode is specific to this regime: math/logic methods feed raw hidden states without head supervision, whereas our PRL objective requires every $\mathbf{h}_k$ to map to the SID vocabulary at every step. STS resolves this incompatibility and is therefore load-bearing in the LLM SID-decoding regime.
We set $N{=}64$ in our implementation. The parameter overhead of \method{} is $\sim$2.5M ($<$0.2\% of a 1.7B LLM).

\subsection{Per-step Reasoning Loss (PRL)}
\label{sec:prl}

Latent reasoning approaches in math and logic domains~\cite{coconut,codi} do not supervise intermediate steps via the prediction head, leaving these steps without direct task-loss gradients. ReaRec's Progressive Reasoning Learning~\cite{rearec} introduces step-wise supervision in the SR-encoder regime, scoring each reasoning state via item-embedding inner products with progressive temperature annealing. PRL adapts this idea to the LLM SID-decoder regime: each $\hk$ is supervised through the LLM's prediction head against the first SID component of the target item, rather than against item embeddings as in~\cite{rearec}. Per-step supervision in this regime is only stable when paired with Soft Token Selection (\S\ref{sec:sts}), since the 14$\times$ norm gap absent in SR encoders would otherwise drive intermediate logits away from the SID vocabulary as $K$ grows.
PRL applies the LLM's prediction head at every step under a logit-temperature schedule:
\begin{equation}
    \mathcal{L}_\text{PRL} = \frac{1}{K}\sum_{k=1}^{K} \text{CE}\bigl(\text{LM\_head}(\hk)\,/\,T_k,\; y\bigr), \quad T_k = T_b + T_s\bigl(1 - \tfrac{k}{K}\bigr)
    \label{eq:prl}
\end{equation}
where $y$ is the first SID component of the target item, and $T_k$ decays linearly from $T_b{+}T_s(1{-}1/K)$ at step $1$ down to $T_b$ at step $K$ (we use $T_b{=}1$, $T_s{=}5$). Higher temperatures at early steps relax the intermediate predictions. The lowest temperature at step $K$ enforces sharp alignment with the target. The total loss is $\mathcal{L} = \mathcal{L}_\text{CE} + \lambda \mathcal{L}_\text{PRL}$ ($\lambda{=}0.1$).
We supervise the first SID token because it carries the coarsest routing signal in the hierarchical RQ-VAE codebook, providing a stable target for intermediate hidden states. Supervising deeper levels would require teacher-forcing the prefix of the target SID at each latent step, which would couple the latent rollout trajectory to a fixed decoding path and conflict with the autoregressive SID decoding used at inference. Later SID levels remain supervised by the standard $\mathcal{L}_\text{CE}$.

PRL trains the Thought Module to produce progressively better latent states under head supervision, and updates the LLM's LoRA adapters through the shared objective. The full training step is summarized in Algorithm~\ref{alg:train}. Empirically, PRL improves performance even at $K{=}0$ inference, where the Thought Module does not participate in the forward pass: the $K{=}0$ gain therefore reflects multi-step PRL supervision shaping the LoRA adapters during training, rather than emergent $K$-step reasoning at inference. PRL can equivalently be read as structured multi-step deep supervision through the LM head. A component-wise ablation (\S\ref{sec:ablation}) shows PRL as the dominant contributor, with the three components compounding to the $+5.0\%$ lift over CoT SFT on Beauty.

\begin{algorithm}[!t]
\caption{\method{} training step (one batch element)}
\label{alg:train}
\small
\begin{algorithmic}[1]
\Require user history $h_\text{user}$, target SID $y=(y^1,\ldots,y^L)$, depth $K_\text{train}$, weight $\lambda$
\State $\hzero \gets$ \Call{LLM}{$h_\text{user}$}.lastHidden \Comment{user representation}
\For{$k=1$ to $K_\text{train}$}
  \State $\mathbf{q}_k \gets W_q\, \mathbf{h}_{k-1}$;\quad $\boldsymbol{\alpha}_k \gets \text{softmax}(\mathbf{q}_k^\top \mathbf{S}/\sqrt{d})$
  \State $\mathbf{e}_k \gets \sum_i \alpha_{k,i}\, \mathbf{s}_i$ \Comment{STS projection}
  \State $\hk \gets \text{LLM}(\mathbf{e}_k, \text{KV}^{(k-1)})$ \Comment{latent token}
  \State $T_k \gets T_b + T_s\,(1 - k/K_\text{train})$
  \State $\ell^{\text{PRL}}_k \gets \text{CE}\bigl(\text{LM\_head}(\hk)/T_k,\, y^1\bigr)$
\EndFor
\State $\ell^{\text{CE}} \gets \sum_{l=1}^{L} \text{CE}\bigl(\text{LM\_head}\circ\text{LLM}(y^{<l},\mathbf{h}_{K_\text{train}}),\, y^l\bigr)$
\State $\mathcal{L} \gets \ell^{\text{CE}} + \lambda \cdot \tfrac{1}{K_\text{train}} \sum_{k=1}^{K_\text{train}} \ell^{\text{PRL}}_k$
\State Update LoRA adapters and $\{\mathbf{S}, W_q\}$ via $\nabla \mathcal{L}$
\end{algorithmic}
\end{algorithm}

\subsection{Training Pipeline}
\label{sec:training}

\method{} starts from a CoT-pre-trained checkpoint, obtained by fine-tuning Qwen3-1.7B on GPT-4-distilled CoT supervision following the recipe of~\cite{onerec-think}. This checkpoint also serves as the CoT SFT baseline (Table~\ref{tab:main}), with the same CoT teacher held constant across both. CoT initialization stabilizes deep latent rollout (\S\ref{sec:analysis}) but is not required for the recommended $K{=}0$ deployment.

\section{Experiments}
\label{sec:exp}

\subsection{Setup}

\noindent\textbf{Datasets and protocol.} Amazon Review 2014~\cite{amazon-review} 5-core: Beauty (22.4K users / 12.1K items), Toys (19.4K / 11.9K), Sports (35.6K / 18.4K). Items are encoded as 4-level RQ-VAE SIDs (codebook 256, 0\% collision after Sinkhorn deduplication)~\cite{tiger}. All methods in Table~\ref{tab:main} share an identical protocol: leave-one-out evaluation, full-corpus ranking, beam width 50, HR@$\{5,10\}$ and NDCG@$\{5,10\}$.

\noindent\textbf{Baselines.} Discriminative references GRU4Rec~\cite{gru4rec}, SASRec~\cite{sasrec}, BERT4Rec~\cite{bert4rec}, HGN~\cite{hgn}; SID-based generative TIGER~\cite{tiger}; latent-reasoning ReaRec~\cite{rearec}; and \emph{LLM-CoT SFT} (henceforth CoT SFT), our same-backbone CoT distillation baseline following the recipe of~\cite{onerec-think}: Qwen3-1.7B fine-tuned on GPT-4-generated reasoning traces (\S\ref{sec:training}), which \method{} directly extends. ReaRec is the only open-source SID-compatible latent-reasoning reference. LARES, PLR (encoder-only) and concurrent LatentR3 (title-decoded LLM) would require porting onto our SID stack and are not directly comparable.

\noindent\textbf{Implementation.} Qwen3-1.7B~\cite{qwen3}, LoRA~\cite{lora} rank 16 applied to all attention and MLP linear layers, $K_\text{train}{=}2$, $\lambda{=}0.1$, PRL temperature $T_b{=}1$, $T_s{=}5$, lr 5e-6 (LoRA) and 5e-5 (Thought Module), batch 64, 3 epochs, 2$\times$H20 GPUs with DeepSpeed ZeRO-2~\cite{deepspeed}. Only LoRA adapters and the 2.5M-parameter Thought Module receive gradients. The LLM backbone is otherwise frozen. RQ-VAE tokenizers are trained per-domain for 100 epochs with Sinkhorn-balanced clustering. Latency in Table~\ref{tab:efficiency} is measured at batch=1 on a single H20 GPU over 100 randomly-sampled test users.

\subsection{Main Results}

\method{} K=0 improves HR@10 by $5.0\%/4.5\%/4.0\%$ over the LLM-CoT SFT baseline (Beauty 0.0625$\to$0.0656, Toys 0.0639$\to$0.0668, Sports 0.0297$\to$0.0309). K=1 yields no consistent gain across datasets ($+0.6\%$/$-0.6\%$/$-0.6\%$), reinforcing $K{=}0$ as the recommended default. The smaller Sports margin tracks a weaker LLM-CoT SFT lead over GRU4Rec/HGN there: language-derived item priors carry less signal on Sports than on Beauty/Toys. Over the strongest discriminative baseline (HGN), \method{} K=0 retains a $+4.1\%$/$+2.9\%$/$+1.3\%$ HR@10 margin. Gains hold across all four metrics (HR@$\{5,10\}$, NDCG@$\{5,10\}$) on every dataset, with no cell where \method{} K=0 underperforms CoT SFT (Table~\ref{tab:main}).
\method{} K=0 also improves over ReaRec by $16.5\%/5.7\%/8.4\%$ HR@10, but ReaRec is an SR-encoder at much smaller parameter scale and the margin partly reflects backbone capacity. CoT SFT alone (Beauty 0.0625) does not surpass HGN (0.0630), while \method{} K=0 (0.0656) does: the LLM advantage in this regime requires training-time supervision.

\begin{table*}[!t]
    \caption{Overall performance on three Amazon benchmarks.}
    \label{tab:main}
    \small
    \setlength{\tabcolsep}{3pt}
    \begin{tabular}{l cccc cccc cccc}
        \toprule
        & \multicolumn{4}{c}{\textbf{Amazon Beauty}} & \multicolumn{4}{c}{\textbf{Amazon Toys}} & \multicolumn{4}{c}{\textbf{Amazon Sports}} \\
        \cmidrule(lr){2-5} \cmidrule(lr){6-9} \cmidrule(lr){10-13}
        \textbf{Method} & HR@5 & HR@10 & N@5 & N@10 & HR@5 & HR@10 & N@5 & N@10 & HR@5 & HR@10 & N@5 & N@10 \\
        \midrule
        \multicolumn{13}{l}{\textit{Traditional Sequential Recommendation}} \\
        GRU4Rec    & 0.0395 & 0.0613 & 0.0260 & 0.0330 & 0.0341 & 0.0492 & 0.0240 & 0.0289 & 0.0197 & 0.0298 & 0.0126 & 0.0164 \\
        SASRec     & 0.0349 & 0.0541 & 0.0202 & 0.0264 & 0.0380 & 0.0609 & 0.0202 & 0.0276 & 0.0162 & 0.0272 & 0.0086 & 0.0121 \\
        BERT4Rec   & 0.0197 & 0.0337 & 0.0121 & 0.0166 & 0.0133 & 0.0225 & 0.0084 & 0.0113 & 0.0095 & 0.0158 & 0.0058 & 0.0078 \\
        HGN        & 0.0420 & 0.0630 & 0.0289 & 0.0356 & 0.0427 & 0.0649 & 0.0287 & 0.0359 & 0.0194 & 0.0305 & 0.0123 & 0.0162 \\
        \midrule
        \multicolumn{13}{l}{\textit{SID-Based Generative Recommendation}} \\
        TIGER                & 0.0346 & 0.0565 & 0.0228 & 0.0299 & 0.0304 & 0.0496 & 0.0200 & 0.0261 & 0.0203 & 0.0302 & 0.0128 & 0.0166 \\
        ReaRec    & 0.0413 & 0.0563 & 0.0244 & 0.0292 & 0.0396 & 0.0632 & 0.0217 & 0.0293 & 0.0154 & 0.0285 & 0.0089 & 0.0131 \\
        \midrule
        \multicolumn{13}{l}{\textit{LLM-Based Methods (Qwen3-1.7B backbone, full-corpus ranking, beam=50)}} \\
        LLM-CoT SFT          & 0.0449 & 0.0625 & 0.0312 & 0.0369 & 0.0453 & 0.0639 & 0.0316 & 0.0376 & 0.0197 & 0.0297 & 0.0132 & 0.0163 \\
        \rowcolor{methodrow} \method{} K=0        & 0.0481 & 0.0656 & 0.0341 & 0.0398 & 0.0482 & 0.0668 & 0.0343 & 0.0403 & 0.0208 & 0.0309 & 0.0142 & 0.0174 \\
        \rowcolor{methodrow} \method{} K=1        & 0.0477 & 0.0660 & 0.0341 & 0.0400 & 0.0483 & 0.0664 & 0.0344 & 0.0402 & 0.0206 & 0.0307 & 0.0140 & 0.0172 \\
        \bottomrule
    \end{tabular}
\end{table*}

\begin{table*}[!t]
\begin{minipage}[t]{0.48\textwidth}
    \centering
    \captionof{table}{Effect of CoT initialization on Beauty. Both bases converge at K=0. Divergence is visible only for deeper rollouts. The $\Delta$ row reports between-base differences at each $K_\text{eval}$.}
    \label{tab:init}
    \small
    \setlength{\tabcolsep}{4pt}
    \begin{tabular}{l c c c c c c}
        \toprule
        \textbf{Base Model} & \textbf{K=0} & \textbf{K=1} & \textbf{K=2} & \textbf{K=4} & \textbf{K=8} & \textbf{K=16} \\
        \midrule
        w/ CoT init  & 0.0656 & 0.0660 & 0.0656 & 0.0650 & 0.0634 & 0.0634 \\
        w/o CoT init & 0.0656 & 0.0644 & 0.0639 & 0.0623 & 0.0602 & 0.0583 \\
        \midrule
        $\Delta$ & 0 & +0.0016 & +0.0017 & +0.0027 & +0.0032 & +0.0051 \\
        \bottomrule
    \end{tabular}
\end{minipage}\hfill
\begin{minipage}[t]{0.48\textwidth}
    \centering
    \captionof{table}{Reasoning cost on Beauty (batch=1, time per sample). $P$ denotes one LLM forward pass.}
    \label{tab:efficiency}
    \small
    \setlength{\tabcolsep}{3pt}
    \begin{tabular}{l c c c c}
        \toprule
        \textbf{Method} & \textbf{Output} & \textbf{Time (s)} & \textbf{Extra FLOPs} & \textbf{HR@10} \\
        \midrule
        LLM-CoT SFT         & 128 tokens & 10.80         & $\sim$256$P$ & 0.0625 \\
        \method{} K=0       & none       & \textbf{0.59} & 0            & 0.0656 \\
        \method{} K=1       & 1 step     & \textbf{0.61} & $\sim$2$P$   & 0.0660 \\
        \bottomrule
    \end{tabular}
\end{minipage}
\end{table*}

\subsection{Ablation Study}
\label{sec:ablation}

PRL supervision is the dominant contributor: it lifts HR@10 from 0.0633 (continued LoRA + Thought Module without PRL) to 0.0656 ($+3.63\%$). Two controls on the same CoT SFT checkpoint with matched epochs, lr, batch, and data isolate this contribution: (i) continued LoRA fine-tuning with SID CE only reaches 0.0629 ($+0.64\%$); (ii) adding the Thought Module and Soft Token Selection without PRL reaches 0.0633 ($+0.64\%$ over (i)); (iii) adding PRL lifts to 0.0656 ($+3.63\%$ increment, Figure~\ref{fig:results} left). The three components compound multiplicatively to the $+5.0\%$ end-to-end lift ($1.0064 \times 1.0064 \times 1.0363 \approx 1.0496$). Control (i) already supervises the first SID token via standard SID CE, so PRL's $+3.63\%$ adds value beyond first-token classification. \S\ref{sec:analysis} further shows the same PRL on a non-CoT base degrades at every $K{>}0$ ($1.8\%$ at K=1, $11.1\%$ at K=16): CoT installs the representational substrate that PRL exploits.

\subsection{Model Analysis}
\label{sec:analysis}

\noindent\textbf{CoT initialization is required for latent rollout.}
Table~\ref{tab:init} reports a controlled reversal: identical PRL training is applied to two checkpoints differing only in pretraining (CoT SFT vs direct SID prediction). Both reach 0.0656 at K=0, but diverge as rollout deepens: the no-CoT base loses $2.6\%/5.0\%/8.2\%/11.1\%$ at K=2/4/8/16, while the CoT base stays within $3.4\%$. CoT supervision is therefore not required for K=0 deployment but is the substrate that keeps deep latent rollout stable.

\noindent\textbf{Inference cost.}
Table~\ref{tab:efficiency} summarizes cost-accuracy on Beauty. \method{} K=0 adds no extra FLOPs over a SID-only decoder yet improves over CoT SFT by $5.0\%$. K=1 adds one single-token LLM forward ($\sim$2$P$ FLOPs) and reaches 0.0660, with no consistent K=1 gain across datasets (Table~\ref{tab:main}). \method{} K=0 reaches $0.59$\,s/sample vs $10.80$\,s for CoT SFT, an $18.3\times$ latency reduction (K=1: $0.61$\,s, $17.7\times$). The reasoning cost moves from inference to training, paid once rather than at every recommendation.
\begin{figure}[t]
    \centering
    \includegraphics[width=\linewidth]{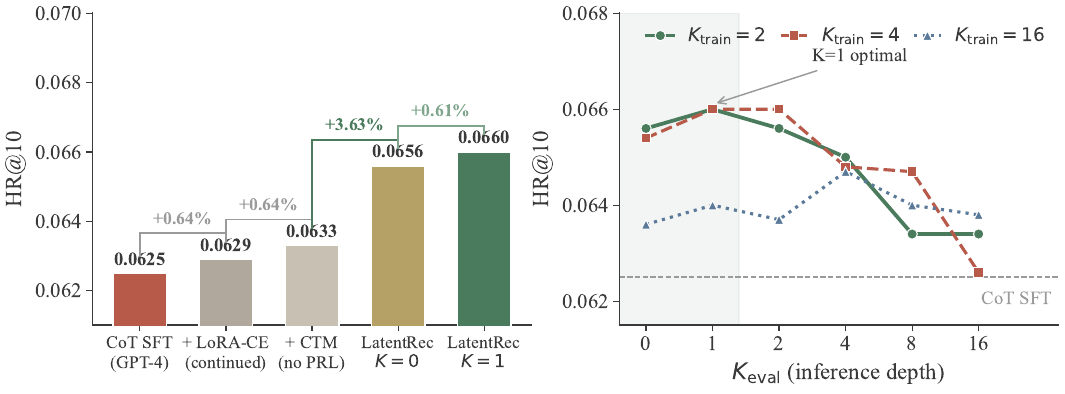}
    \caption{\textbf{(a)} Component-wise decomposition of \method{}'s gain on Beauty HR@10 at $K{=}0$. \textbf{(b)} HR@10 across $K_\text{eval}$ for $K_\text{train}\in\{2,4,16\}$. Dashed line marks the CoT SFT baseline.}
    \Description{Two side-by-side panels. Left: 5-bar ablation chart on Beauty HR@10. Right: line chart of K-sweep across training depths.}
    \label{fig:results}
\end{figure}

\noindent\textbf{Reasoning depth.}
The useful $K_\text{eval}$ range scales with $K_\text{train}$, bounding when latent rollout helps and when it hurts. Figure~\ref{fig:results}b shows the full $K_\text{train} \times K_\text{eval}$ sweep on Beauty. Three findings emerge: (1)~$K_\text{train}{=}2$ yields the strongest K=0 base (0.0656 vs.\ 0.0636 for $K_\text{train}{=}16$), as shorter chains concentrate PRL supervision closer to the final prediction; (2)~K=1 is optimal across all training depths, with deeper rollouts plateauing or degrading; (3)~the useful $K_\text{eval}$ range scales roughly with $K_\text{train}$: for $K_\text{train}{=}2$ only K=1 improves, while for $K_\text{train}{=}16$ improvement extends to K=4.

\section{Conclusion}

\method{} demonstrates that inference-time reasoning is not a prerequisite for LLM-based generative recommendation: per-step prediction-head supervision during training improves HR@10 by $4.0$--$5.0\%$ over a CoT-distilled baseline while reducing inference latency by $18.3\times$. Two findings clarify CoT's role in this regime. (1) K$\geq$1 latent rollout yields no consistent benefit across datasets, locating the value of multi-step latent computation in training rather than inference. (2) CoT pre-training stabilizes deep latent rollout but is not required for K=0, separating warm-start from per-step supervision as distinct contributors. Training-time supervision, not inference-time scaling, is the more productive lever for reasoning in LLM-based recommendation.

\noindent\textbf{Limitations and Future Work.} Our empirical scope is a single LLM backbone (Qwen3-1.7B), a single CoT teacher (GPT-4), and three Amazon e-commerce domains. Cross-backbone, cross-teacher, and cross-domain generalization (e.g., movies, news, music) remains open, as does extension to non-SID generative recommenders. A natural next step is supervising deeper SID levels without coupling the latent rollout to a fixed decoding path.

\section*{GenAI Usage Disclosure}

We used GPT-4 to generate chain-of-thought reasoning text for the LLM-CoT SFT baseline that \method{} extends. Each generation is conditioned on the user's purchase history and the last 2--3 levels of the target item's category path. The target's title and Semantic ID are withheld, and the full prompt template will be released with the code upon acceptance. We additionally used a generative AI assistant for minor grammar checks during writing.

\bibliographystyle{ACM-Reference-Format}
\balance
\bibliography{references}

\end{document}